\pdfoutput=1

\documentclass[11pt]{article}

\usepackage[final]{acl}

\usepackage{times}
\usepackage{latexsym}

\usepackage[T1]{fontenc}

\usepackage[utf8]{inputenc}

\usepackage{microtype}

\usepackage{inconsolata}

\usepackage{graphicx}
\usepackage{subcaption}

\usepackage{paralist}
\usepackage{xspace}

\usepackage{tabularx}
\usepackage{booktabs}
\usepackage{multirow}
\usepackage{pifont}
\usepackage{threeparttable}
\usepackage{amsmath}

\DeclareMathOperator*{\argmax}{arg\,max}
\DeclareMathOperator*{\argmin}{arg\,min}

\usepackage[capitalize]{cleveref} 

\crefformat{subfigure}{#2Figure~#1#3}

\newcommand{\method}{\texttt{EnAnchored-X2X}\xspace}

\title{\method: English-Anchored Optimization for Many-to-Many Translation}

\author{
 \textbf{Sen Yang\textsuperscript{1}},
 \textbf{Yu Bao\textsuperscript{2}},
 \textbf{Yu Lu\textsuperscript{2}},
 \textbf{Jiajun Chen\textsuperscript{1}},
 \textbf{Shujian Huang\textsuperscript{1*}},
 \textbf{Shanbo Cheng\textsuperscript{1,2*}}
\\
 \textsuperscript{1}National Key Laboratory for Novel Software Technology, Nanjing University,
 \\
 \textsuperscript{2}ByteDance Research,
\\
\small{
  \texttt{yangsen@smail.nju.edu.cn},\quad\texttt{\{huangsj,chenjj\}@nju.edu.cn}
}\\
\small{
  \texttt{\{baoyu.001,luyu.ly,chengshanbo\}@bytedance.com}
}
}

\begin{document}
\maketitle
\renewcommand{\thefootnote}{\fnsymbol{footnote}}
\footnotetext[1]{Corresponding author}
\renewcommand{\thefootnote}{\arabic{footnote}}

\begin{abstract}

Large language models (LLMs) have demonstrated strong machine translation capabilities for English-centric language pairs but underperform in direct non-English (x2x) translation. This work addresses this limitation through a synthetic data generation framework that leverages models' established English-to-x (en2x) capabilities. By extending English parallel corpora into omnidirectional datasets and developing an English-referenced quality evaluation proxy, we enable effective collection of high-quality x2x training data. Combined with preference-based optimization, our method achieves significant improvement across 72 x2x directions for widely used LLMs, while generalizing to enhance en2x performance. The results demonstrate that strategic exploitation of English-centric strengths can bootstrap comprehensive multilingual translation capabilities in LLMs. We release codes, datasets, and model checkpoints at \url{https://github.com/NJUNLP/EAX}

\end{abstract}

\section{Introduction}

\begin{figure}[tb]
    \centering
    \includegraphics[scale=0.375]{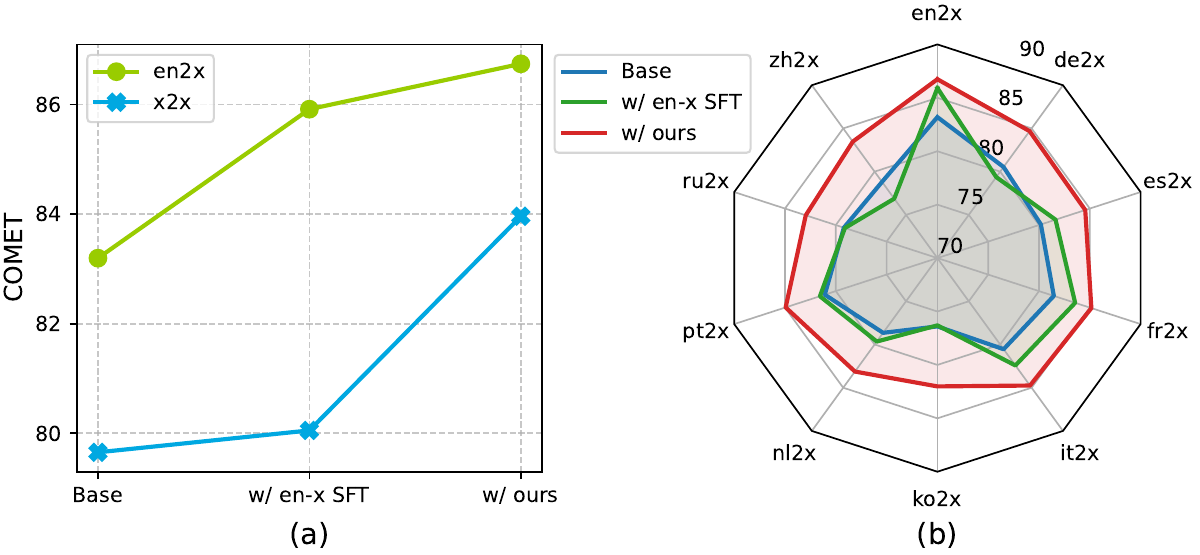}
    \vspace{-2ex}
   \caption{COMET score of the Llama2 base model on en2x and x2x language pairs with en-x supervised fine-tuning and our x2x optimization (a), as well as performance based on source language categorization~(b).}
    \label{fig:perf_overview}
\end{figure}

Recent advances in large language models (LLMs) have propelled significant progress in machine translation~\cite{alves2024tower,xu2023paradigm}.
This is largely attributed to the incorporation of multilingual data alongside predominantly English data during pre-training, enabling models to develop multilingual capabilities.
While LLMs can typically achieve competent translation abilities between English and other languages through fine-tuning with minimal parallel data, we observe that these translation capabilities do not generalize effectively across non-English language pairs.
Specifically, direct translation capabilities between non-English languages (x2x) substantially lag behind their performance in English-centric translation (en2x), as illustrated in Figure~\ref{fig:perf_overview}.
Despite the critical importance for real-world applications requiring multilingual communication beyond just English.
While using English as a pivot language offers a compromise solution, this approach often suffers from error propagation and doubles the decoding overhead compared to direct translation, motivating our exploration of methods to enhance models' omnidirectional translation capabilities.

A straightforward approach to improving models' translation capabilities between non-English languages would be to collect high-quality parallel corpora for fine-tuning, similar to how we enhance English translation capabilities. However, non-English language parallel data is scarce and challenging to scale. This limitation stems from the prohibitive costs of annotation in non-English language directions (faced with a shortage of qualified expert translators) and the quadratic growth in language pairs as the number of languages increases.

Synthetic data has emerged as a promising alternative to annotated corpora for enhancing multilingual capabilities, with recent advancements demonstrating its scalability and potential to augment various LLM functionalities~\cite{long-etal-2024-llms, yu2023large, huang-etal-2023-large}.
However, generating high-quality non-English parallel corpora for translation tasks via LLMs remains nontrivial, fundamentally constrained by two interconnected challenges:
\begin{compactitem}
    \item Direct cross-lingual generation (x2x) between low-resource languages suffers from LLMs' limited native translation expertise, leading to outputs with unsatisfactory quality.
    \item Synthetic data inherently lacks built-in quality guarantees, necessitating rigorous curation. Yet, unlike English-centric tasks, x2x translation lacks reliable automatic evaluation metrics, making data filtering both critical and methodologically underspecified.
\end{compactitem}

To address these challenges, we propose our method, \method, which leverages the en2x capabilities of LLMs and abundant English parallel corpora. 
First, we extend existing English parallel data into an omnidirectional dataset through synthesis. At the generation, we provide the model with both the source language text and its English reference, effectively giving the model two source texts (one being English) before requesting translation into another language. This approach allows the model to utilize its en2x capabilities during translation, resulting in higher quality outputs.

Second, we develop an en2x evaluation model using existing en-x parallel data and adapt it for x2x assessment by transforming x2x evaluation into en2x evaluation. Specifically, we substitute the source text with its English reference and use the model to evaluate the score between this English reference and the target text as a proxy for the original translation quality assessment.

Finally, integrating our translation synthesis and evaluation strategies enables the collection of high-quality x2x data. To further exploit the potential of synthetic data, we retain lower-quality translations to create preference pairs with high-quality translations, enabling preference-based optimization of the model.

We apply our methodology across three distinct base models and observe comprehensive improvements in x2x translation capabilities, exemplified by an average increase of 7 points in BLEURT scores across 72 x2x language pairs for the Llama2 model. A particularly intriguing finding is the sustained enhancement in en2x translation performance, even though these language pairs are outside our optimization scope. Our investigation into different optimization algorithms reveals that our approach demonstrates increasingly significant benefits with data scaling and exhibits robust generalization of translation capabilities across diverse linguistic contexts.

\section{Generalization of Non-English Language Translation}

To examine the generalization of existing models across non-English language pairs, we first conducted supervised fine-tuning~(SFT) using widely available parallel corpora. 
Given the predominant English-centric alignment in existing multilingual datasets, the models demonstrated predictable robustness in English-centric~(en2x/x2en) directions. 
However, our investigation focused on a critical yet understudied phenomenon: whether cross-lingual transfer between non-English languages~(x2x) could emerge from such English-anchored training paradigms.

We utilize TowerBlocks~\cite{alves2024tower}, encompassing parallel data between English and nine languages, approximately 150k samples in total. Figure~\ref{fig:perf_overview} demonstrates the en2x and x2x performance of the Llama2 base model~\cite{touvron2023llama} after SFT on the translation data. While the model shows marked improvement in en2x performance post-fine-tuning, the x2x performance presents a more complex picture: only three languages (Spanish, French, and Italian) exhibit significant improvement, while the remaining languages show negligible performance changes. We even observe performance degradation in zh2x and de2x directions. Overall, the SFT process leads to a widening performance gap between en2x and x2x translations, suggesting that the model's translation capabilities between multiple languages are not fully activated under the current training setup.

\section{Methodology}

\begin{figure*}[tb]
    \centering
    \includegraphics[width=\textwidth]{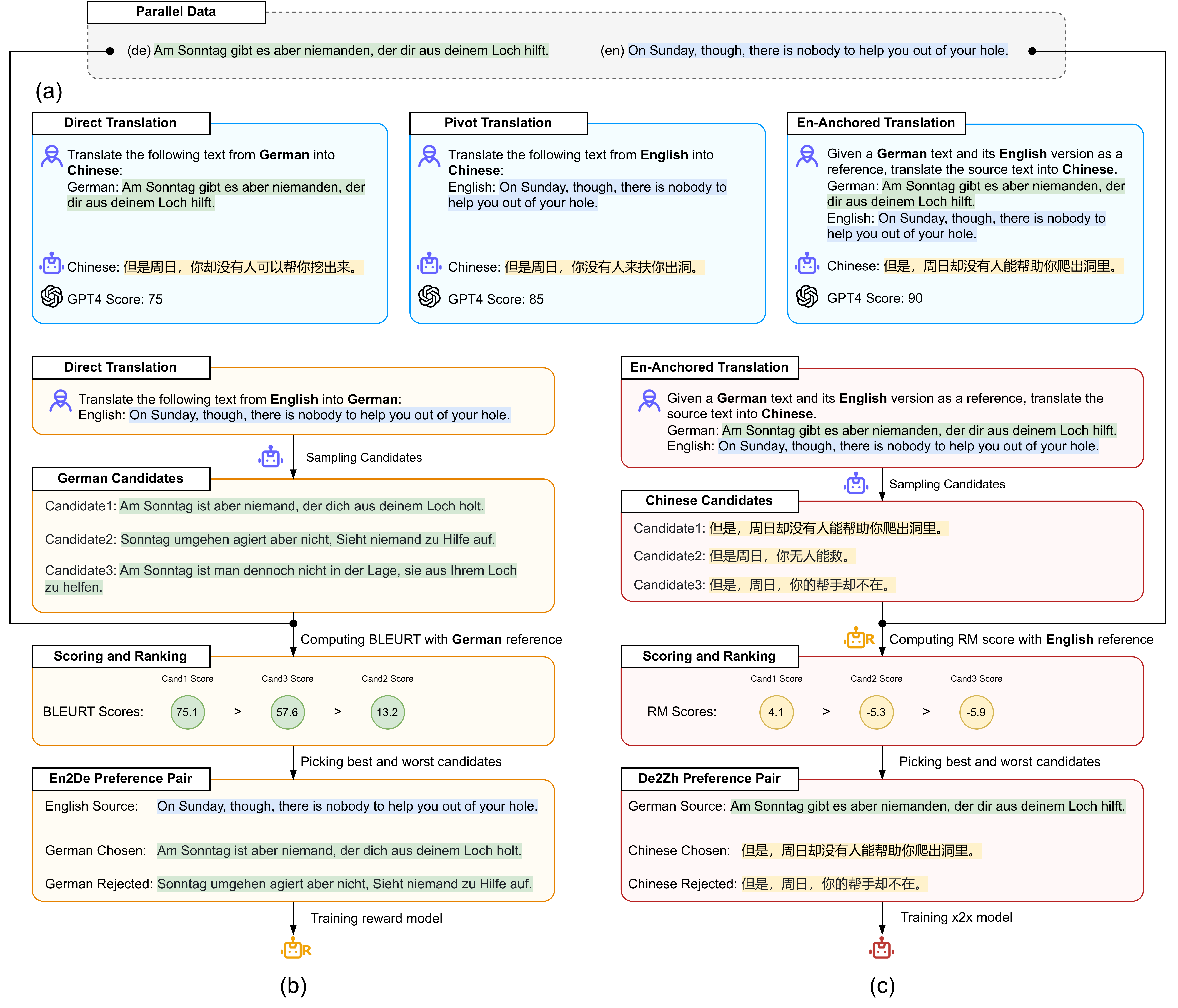}
    \vspace{-2ex}
   \caption{The overview of \method. Based on existing parallel data, the comparison of three methods for synthesizing x2x translation data (a), the process of constructing reward model for en2x evaluation (b) and the x2x preference data construction (c).}
    \label{fig:arch_overview}
\end{figure*}

To address the model's generalization deficiencies between non-English languages, there is an urgent need to enrich the diversity of language pairs in existing training data by extending current English-centric parallel data to cover all language directions. Our data synthesis pipeline comprises three components: Section~\ref{sec:eaxt} introduces our data synthesis method based on English-Anchored translation, Section~\ref{sec:reward_model} presents our English-Anchored data evaluation framework, and Section~\ref{sec:construct_preference_data} details our process for data selection and preference pair construction. All these components leverage the LLM's inherent capabilities and existing parallel data.

\subsection{English-Anchored x2x Translation (EAxT)}
\label{sec:eaxt}

\begin{table}[tbp]
\small
\centering
\begin{tabular}{lc}
\toprule
\textbf{Synthetic Strategy} & {\textbf{GPT4 Score}}\\
\midrule
Direct      & 78.46 (22.43)  \\
Pivot      & 79.58 (18.89)  \\
EAxT      & 82.26 (18.41) \\
\bottomrule
\end{tabular}
\caption{The GPT4 quality scores~\cite{kocmi-federmann-2023-large} of the translations for synthetic methods, along with the standard deviations are in parentheses.}
\label{tab:gpt_score}
\end{table}

Given a parallel data pair $(x_{l_1},x_{\text{en}})\in D$, where $x_{l_1}$ represents source text in language $l_1$ and $x_{\text{en}}$ its English annotation, and targeting language $l_2$, let $\bar{x}_{l_2}^{\text{direct}}\sim M(x|x_{l_1})$ denote the model's direct translation. Previous experiments have demonstrated that direct x2x translation suffers from quality deficiencies. Conversely, considering the model's superior performance in en2x translation, leveraging this capability to generate data for x2x optimization appears promising. One approach involves utilizing pivot translation, generating $\bar{x}_{l_2}^{\text{pivot}}\sim M(x|x_{\text{en}})$ by translating through English. However, this inherits pivot translation's drawbacks: lack of direct alignment between pivot-translated text and source text, and risk of error propagation.

We propose combining direct and pivot translation to obtain higher quality translation data through English-Anchored x2x Translation (EAxT). Specifically, we simultaneously provide the model with both the non-English source text $x_{l_1}$ and its English translation $x_{\text{en}}$ as reference, then request translation into target language $l_2$, i.e., $\bar{x}_{l_2}^{\text{EAxT}}\sim M(x|x_{l_1},x_{\text{en}})$. During this process, the model's access to the English reference enables flexible integration of its en2x translation capabilities into the x2x translation process. As illustrated in Figure~\ref{fig:arch_overview} (a), we find that LLMs can excel at this task without additional training, thanks to their robust comprehension and instruction-following capabilities.

We sampled 7,200 instances (100 per language pair) and compared the quality of translations generated by these three synthesis methods. Lacking human-annotated reference translations, we employed GPT-4 to evaluate the quality of the model-generated x2x translations. Results are presented in Table~\ref{tab:gpt_score} and the box plot in Figure~\ref{fig:gpt_score} further illustrates the distribution.
The results demonstrate that EAxT-generated data achieves higher quality on average compared to other methods. Moreover, we observed substantial score variations at the sample level, indicating instability in synthetic data quality across different samples, necessitating large-scale evaluation and filtering.

\subsection{English-Anchored x2x Evaluation (EAxE)}
\label{sec:reward_model}

Without careful design and validation, synthetic data may amplify existing biases, introduce new ones, or even trigger model collapse~\cite{seddik2024bad}. A common challenge in large-scale synthetic data application is ensuring the factuality and fidelity~\cite{liu2024best}. For translation tasks, without proper evaluation and filtering of synthetic translations, we cannot provide clear guidance for model optimization, thereby limiting the ultimate performance ceiling.

Obtaining evaluation scores directly for x2x directions is a non-trivial problem, so we consider converting x2x evaluation into en2x evaluation. Ideally, $s(x_{l_1},\bar{x}_{l_2})$ represents the quality score between the source text $x_{l_1}$ and the generated translation $\bar{x}_{l_2}$, measuring their alignment. Since the English reference $x_{\text{en}}$ for source text $x_{l_1}$ is accessible, we can assume that the semantic consistency between $x_{\text{en}}$ and translation $\bar{x}_{l_2}$ correlates positively with the consistency between $x_{l_1}$ and $\bar{x}_{l_2}$, i.e.,
\begin{equation}
\label{equ:score_correlation}
s(x_{l_1},\bar{x}_{l_2})\propto s(x_{\text{en}},\bar{x}_{l_2}).
\end{equation}

Thus, by using $s(x_{\text{en}},\bar{x}_{l_2})$ as a proxy for $s(x_{l_1},\bar{x}_{l_2})$, we convert x2x evaluation into en2x evaluation. Evaluating en2x translation quality is relatively straightforward, as models already possess strong en2x translation capabilities after SFT, suggesting its potential for en2x evaluation task. Furthermore, we can again leverage existing en2x parallel corpora to activate the model's en2x evaluation capabilities.

To enable models to assess translation quality and output a score, we implement this idea through Reward Modeling, a crucial component widely applied in reinforcement learning process. Its primary task is to predict reward values based on given inputs, thereby guiding the direction of the learning algorithm. For translation task, this reward value can be considered as a score of quality.

Training a reward model requires a preference dataset. For en2x evaluation~(e.g., English to language $l_1$), we need to collect preference pairs comprising a good and a bad translation in language $l_1$ for each English source text, denoted as~$(x_{\text{en}},\bar{x}_{l_1}^{\text{chosen}},\bar{x}_{l_1}^{\text{rejected}})$.

Based on existing parallel data $(x_{l_1},x_{\text{en}})\in D$, we provide $x_{\text{en}}$ as source text to the model, requesting translations to language $l_1$ and sampling $n$ results $\bar{x}_{l_1}^i\sim M(x|x_{\text{en}})$, where $i\in[n]$. The key difference with the x2x evaluation is that we can compute quality scores $s(x_{\text{en}},\bar{x}_{l_1}^i)=\text{bleurt}(x_{l_1},\bar{x}_{l_1}^i)$ for each translation $\bar{x}_{l_1}^i$ using the annotated reference translation $x_{l_1}$. Finally, we select the best and worst translations from the samples to form preference pairs:
\begin{equation}
\begin{split}
    \bar{x}_{l_1}^{\text{chosen}} &= \argmax_{\bar{x}_{l_1}^i}{s(x_{\text{en}},\bar{x}_{l_1}^i)}, \\
    \bar{x}_{l_1}^{\text{rejected}} &= \argmin_{\bar{x}_{l_1}^i}{s(x_{\text{en}},\bar{x}_{l_1}^i)}.
\end{split}
\end{equation}

To train with the preference data, the model is required to score each preference pair, and a Ranking Loss function is employed for optimization, aiming to maximize the score margin between chosen and rejected samples. The complete process is illustrated in Figure~\ref{fig:arch_overview} (b).

\subsection{Preference Data Construction}
\label{sec:construct_preference_data}

In this section, we apply our advanced data synthesis strategy and evaluation method to all possible language pairs to construct large-scale, high-quality x2x translation data.

For a given language pair $l_1\rightarrow l_2$, and parallel data $(x_{l_1},x_{\text{en}})\in D$ as the source, we utilize the EAxT technique introduced in Section~\ref{sec:eaxt} to sample a batch of candidate translations in the target language: $\bar{x}_{l_2}^i\sim M(x|x_{l_1},x_{\text{en}}),\ i\in[n]$. These candidates are then scored using the reward model constructed in Section~\ref{sec:reward_model}. According to Eq.~\ref{equ:score_correlation}, the quality score $s^i$ for candidate $\bar{x}_{l_2}^i$ can be approximated using its score with $x_{\text{en}}$ as a proxy, i.e., $s^i=r(x_{\text{en}},\bar{x}_{l_2}^i)$, where $r(\cdot,\cdot)$ is the translation quality score estimated using the reward model.

Now, with a clear landscape of the data quality, we can proceed with constructing training data. At its simplest, we can retain the highest-scoring candidate $\bar{x}_{l_2}^{\text{chosen}}=\bar{x}_{l_2}^{\argmax_{i}{s^i}}$ to form parallel data $(x_{l_1},\bar{x}_{l_2}^{\text{chosen}})$ in pair $l_1\rightarrow l_2$ for fine-tuning. To more effectively utilize synthetic data, we suggest additionally retaining the lowest-scoring candidate $\bar{x}_{l_2}^{\text{rejected}}=\bar{x}_{l_2}^{\argmin_{i}{s^i}}$, creating preference data $(x_{l_1},\bar{x}_{l_2}^{\text{chosen}},\bar{x}_{l_2}^{\text{rejected}})$, which provides clearer signals for x2x optimization. Furthermore, preference confidence can be measured by the score margin $s^{\text{chosen}}-s^{\text{rejected}}$. By discarding samples with low confidence, we can control the preference accuracy of data.

Based on the collected preference data, we perform Direct Preference Optimization~\cite[DPO,][]{rafailov2023direct} training for the model. 
This technique has been widely applied across various tasks and has demonstrated superior generalization compared to SFT.

\section{Experiments}

\subsection{Experiment Settings}
To systematically validate the effectiveness and generalizability of our x2x translation framework, we design experiments following a structured pipeline: defining the task scope, selecting representative models, preparing synthetic datasets, and establishing comparative baselines. Below is the detailed setup.

\paragraph{Task.} 
Our primary focus is on 72 cross-lingual~(x2x) translation directions testsets from the FLORES-200 benchmark~\cite{nllb2022flores}, which includes nine representative languages: German (de), French (fr), Dutch (nl), Italian (it), Spanish (es), Portuguese (pt), Korean (ko), Russian (ru), and Chinese (zh). 
This set includes intra-family scenarios (e.g., de$\to$fr within Indo-European) and cross-family cases (e.g., zh$\to$ru between Sino-Tibetan and Slavic). 
We also evaluate x2en~(non-English$\to$English) and en2x~(English$\to$non-English) directions to analyze cross-lingual knowledge spillover from x2x optimization.
For high-resource validation, we supplement with WMT22 de2fr and fr2de~\cite{kocmi-etal-2022-findings} test sets.

\paragraph{Metrics.} 
Translation quality is mainly measured using two metrics: COMET-22~\cite{rei-etal-2022-comet}, a neural metric trained on human preferences to assess semantic adequacy and fluency; and BLEURT-20~\cite{sellam-etal-2020-bleurt}, a reference-based metric optimized for low-resource languages.
Additionally, we employ the GEMBA-SQM template~\cite{kocmi-federmann-2023-large} to obtain reference-free quality scores (0-100 scale) from GPT-4, providing an additional perspective that complements the reference-based metrics.

\paragraph{Base Models.}
We instantiate our method on three 7B-parameter models with diverse multilingual baselines: (i) Llama2-7B~\cite{touvron2023llama}, a vanilla open-source LLM; (ii) TowerBase-7B~\cite{alves2024tower}, a Llama2 variant enhanced with 1.2T tokens of multilingual pretraining (monolingual + parallel data) for cross-lingual tasks; and (iii) Qwen2.5-7B~\cite{qwen2025qwen25technicalreport}, a Chinese-optimized model with improved cross-lingual attention for non-Latin scripts.

\paragraph{Seed Datasets and Implementation.}
For synthesizing x2x training data, we utilize a translation task subset from the TowerBlocks collection~\cite{alves2024tower} as our seed corpus. 
This dataset also serves as the foundation for en-x fine-tuning of base models and reward modeling in Section~\ref{sec:reward_model}. 
The seed corpus comprises about 150k parallel sentences covering nine non-English languages. 
For each non-English source text, we generate translations into the other eight languages, yielding approximately 1M data entries. 
We sample four candidate translations per entry and employ our evaluation strategy to score them for constructing preference pairs. 
After filtering out low-confidence preference pairs based on score margins, the final preference data used for training consists of approximately 140k pairs for Llama2, 210k pairs for Qwen2.5 and 250k pairs for Tower.
The training hyperparameters and implementation details are explained in Appendix~\ref{sec:details}.

\paragraph{Baselines.}
We compare against the following baselines representing diverse strategies:
\begin{compactitem}
    \item \textbf{Base Model}~(untuned, 7B parameters): establishes a pretrained performance baseline.
    \item \textbf{SFT Model}: the base model fine-tuned on 150K en-x seed datasets, represents English-centric optimization.
    \item \textbf{FLORES x2x SFT}: the SFT model further fine-tuned on 72K human-annotated x2x pairs with 1K per direction.
    \item \textbf{Pivot Translation}: two-stage translation strategy via English intermediate.
    \item \textbf{TowerInstruct-7B}~\cite{alves2024tower}: This model is fine-tuned from TowerBase using 640k multi-task annotated data, encompassing tasks beyond translation such as paraphrasing, translation quality estimation, and named entity recognition.
    \item \textbf{M2M-100-12B}~\cite{fan2020beyond}: This work constructed an x2x dataset through large-scale mining, including 7.5 billion parallel data entries across 100 languages, resulting in a model capable of translation among 100 languages.
\end{compactitem}

\subsection{Main Results}

\begin{table*}[htbp]
\small
\centering
\begin{tabular}{lcccccccc}
\toprule
\multirow{2}{*}{\textbf{Models}}& \multicolumn{2}{c}{\textbf{x2en}} & \multicolumn{2}{c}{\textbf{en2x}} & \multicolumn{2}{c}{\textbf{x2x}} & \multicolumn{2}{c}{\textbf{AVG}} \\
\cmidrule(lr){2-3} \cmidrule(lr){4-5} \cmidrule(lr){6-7} \cmidrule(lr){8-9}
 & \scriptsize{\textbf{BLEURT}} & \scriptsize{\textbf{COMET}} & \scriptsize{\textbf{BLEURT}} & \scriptsize{\textbf{COMET}} & \scriptsize{\textbf{BLEURT}} & \scriptsize{\textbf{COMET}} & \scriptsize{\textbf{BLEURT}} & \scriptsize{\textbf{COMET}} \\
\midrule
TowerInstruct-7B & 78.29 & 88.28 & 75.98 & 88.44 & 71.80 & 85.68 & 72.87 & 86.22 \\
M2M-100-12B     & 75.44 & 85.86 & 72.61 & 82.90 & 69.57 & 85.03 & 70.46 & 84.90 \\
\midrule
Llama2-7B     & 75.24 & 86.34 & 68.56 & 83.20 & 63.42 & 79.65 & 65.12 & 80.68 \\
Llama2-7B-SFT      & 76.78 & 87.43 & 72.05 & 85.91 & 61.92 & 80.05 & 64.42 & 81.38 \\
\quad w/ Pivot Trans. & - & - & - & - & 68.53 & 83.41 & - & - \\
\quad w/ FLORES x2x SFT & 76.29 & 87.01 & 71.69 & 85.77 & 67.94 & 83.17 & 69.15 & 83.82 \\
\quad w/ \method & \textbf{77.15} & \textbf{87.62} & \textbf{73.06} & \textbf{86.74} & \textbf{68.91} & \textbf{83.96} & \textbf{70.15} & \textbf{84.60} \\
\midrule
TowerBase-7B    & 76.98 & 87.47 & 73.75 & 86.79 & 62.76 & 80.57 & 65.28 & 81.88 \\
TowerBase-7B-SFT & 78.15 & 88.21 & 76.14 & 88.46 & 68.17 & 83.81 & 69.96 & 84.71 \\
\quad w/ Pivot Trans. & - & - & - & - & 72.68 & 86.06 & - & - \\
\quad w/ FLORES x2x SFT & 77.58 & 87.80 & 75.44 & 88.09 & 71.99 & 85.72 & 72.89 & 86.16 \\
\quad w/ \method & \textbf{78.36} & \textbf{88.33} & \textbf{76.73} & \textbf{88.86} & \textbf{72.95} & \textbf{86.30} & \textbf{73.87} & \textbf{86.76} \\
\midrule
Qwen2.5-7B      & 77.48 & 87.80 & 71.93 & 86.02 & 69.28 & 84.13 & 70.37 & 84.69 \\
Qwen2.5-7B-SFT       & 77.75 & 87.96 & 74.00 & 87.23 & 70.20 & 84.72 & 71.34 & 85.29 \\
\quad w/ Pivot Trans. & - & - & - & - & 70.69 & 84.89 & - & - \\
\quad w/ FLORES x2x SFT & 76.51 & 87.16 & 73.50 & 86.91 & 70.07 & 84.60 & 71.06 & 85.08 \\
\quad w/ \method & \textbf{78.01} & \textbf{88.09} & \textbf{74.96} & \textbf{87.87} & \textbf{71.44} & \textbf{85.39} & \textbf{72.44} & \textbf{85.91} \\
\bottomrule
\end{tabular}
\caption{Aggregated performance on FLORES-200 testset across 90 translation directions (9 for x2en, 9 for en2x and 72 for x2x).}
\label{tab:main_flores}
\end{table*}

Table~\ref{tab:main_flores} presents the average performance of our method and baselines on the FLORES-200 test set. We report the improvements on the individual languages in Appendix~\ref{sec:results_in_langs}.

Our x2x optimization framework achieves significant performance uplifts over English-centric baselines. 
For Llama2-7B, the x2x BLEURT score improves from 63.42 (Base) to 68.91~(+5.49), with COMET gains of +4.31 points. 
Notably, the optimized TowerBase-7B surpasses both TowerInstruct-7B (a multi-task fine-tuned model) and M2M-100-12B on x2x tasks, achieving 72.95 BLEURT and 86.30 COMET, demonstrating that our synthetic data pipeline can rival large-scale mined datasets like M2M-100's 7.5B pairs.

Despite focusing solely on x2x optimization~(without direct en2x and x2en supervision), our method induces collateral improvements in English-related directions. 
Specifically, the en2x BLEURT of Llama2-7B and Qwen2.5-7B improves 4.50 and 3.03, and outperforms their SFT counterparts~(+1.01 and +0.96), respectively. 
TowerBase-7B also achieves 76.73 en2x BLEURT~(+0.59 over its SFT version).
This suggests that our x2x optimization fosters a more cohesive multilingual semantic space, where cross-lingual knowledge transfer occurs implicitly through English anchoring.

Fine-tuning on the FLORES devset~(72K x2x pairs) improves x2x performance for most models --- e.g., TowerBase gains +3.82 BLEURT points --- though Qwen shows no benefit. Critically, this comes at the cost of x2en or en2x degradation~(e.g., $-$0.7 BLEURT for en2x on TowerBase).
This is likely due to the low diversity of FLORES data, causing overfitting to specific language pairs.
Detailed analysis is in Section~\ref{sec:scaling}.

Although pivot translation achieves competitive x2x scores on FLORES (Table~\ref{tab:main_flores}), it underperforms our \method on WMT22 de2fr and fr2de (Table~\ref{tab:main_wmt}).
This discrepancy stems from FLORES' annotation biases~\cite{zhang-toral-2019-effect,nllb2022flores}: non-English references are derived from English source texts, giving pivot methods an inherent alignment advantage. 
In contrast, WMT22's bidirectional data requires genuine cross-lingual competence, where our en2x-anchored generation proves more robust.

Table~\ref{tab:main_flores_gpt_score} presents the GPT4 assessment results of our method on the FLORES-200 test set. These results align closely with our reference-based evaluation results. Notably, this reference-free evaluation approach eliminates biases introduced by translationese effects~\cite{nllb2022flores,zhang-toral-2019-effect}, enabling us to compare the performance across different language directions more fairly. Our analysis reveals that performance in x2x directions has substantially improved, nearly matching that of en2x directions. This convergence indicates that our method successfully aligns translation capabilities between English and other languages. However, due to the inherent advantage of English data in pre-training corpora, the generation capabilities for non-English target languages (both en2x and x2x) continue to lag behind those for English targets (x2en).

\begin{table}[tbp]
\small
\tabcolsep 2pt
\centering
\begin{tabular}{lcccc}
\toprule
\multirow{2}{*}{\textbf{Models}}& \multicolumn{2}{c}{\textbf{de2fr}} & \multicolumn{2}{c}{\textbf{fr2de}} \\
\cmidrule(lr){2-3} \cmidrule(lr){4-5}
 & \scriptsize{\textbf{BLEURT}} & \scriptsize{\textbf{COMET}} & \scriptsize{\textbf{BLEURT}} & \scriptsize{\textbf{COMET}}\\
\midrule
Llama2-7B-SFT      & 64.19 & 79.33 & 72.08 & 82.07 \\
\quad w/ Pivot Trans. & 64.91 & 79.76 & 72.34 & 82.10 \\
\quad w/ \method & \textbf{65.85} & \textbf{80.18} & \textbf{73.80} & \textbf{83.27} \\
\midrule
TowerBase-7B-SFT      & 69.89 & 82.46 & 76.29 & 85.53 \\
\quad w/ Pivot Trans. & 70.13 & 82.70 & 76.63 & 85.48 \\
\quad w/ \method & \textbf{71.20} & \textbf{83.23} & \textbf{77.57} & \textbf{86.25} \\
\midrule
Qwen2.5-7B-SFT       & 67.53 & 81.34 & 73.47 & 83.41 \\
\quad w/ Pivot Trans. & 68.04 & 81.37 & 74.30 & 83.60 \\
\quad w/ \method & \textbf{69.18} & \textbf{82.20} & \textbf{74.96} & \textbf{84.09} \\
\bottomrule
\end{tabular}
\caption{Performance on the WMT22 de-fr testset.}
\label{tab:main_wmt}
\end{table}

\begin{table}[htbp]
\small
\centering
\tabcolsep 5pt
\begin{tabular}{lcccc}
\toprule
\textbf{Models} & \textbf{x2en} & \textbf{en2x} & \textbf{x2x} & \textbf{AVG} \\
\midrule
Llama2-7B-SFT      & 90.22 & 82.30 & 77.85 & 79.54 \\
\quad w/ \method & \textbf{90.32} & \textbf{83.94} & \textbf{82.51} & \textbf{83.43} \\
\midrule
TowerBase-7B-SFT & \textbf{91.94} & 89.41 & 86.69 & 87.48 \\
\quad w/ \method & 91.88 & \textbf{89.92} & \textbf{89.19} & \textbf{89.53} \\
\midrule
Qwen2.5-7B-SFT       & 91.94 & 85.99 & 86.37 & 86.89 \\
\quad w/ \method & \textbf{92.00} & \textbf{87.16} & \textbf{86.93} & \textbf{87.46} \\
\bottomrule
\end{tabular}
\caption{Aggregated GPT4 assessment~\cite{kocmi-federmann-2023-large} on FLORES-200 testset across 90 translation directions (9 for x2en, 9 for en2x and 72 for x2x).}
\label{tab:main_flores_gpt_score}
\end{table}

\subsection{Ablation Study}

\begin{table*}[htbp]
\small
\tabcolsep 3pt
\centering
\begin{tabular}{llllllllll}
\toprule
 \multicolumn{2}{c}{\multirow{2}{*}{\textbf{Models}}} & \multicolumn{2}{c}{\textbf{x2en}} & \multicolumn{2}{c}{\textbf{en2x}} & \multicolumn{2}{c}{\textbf{x2x}} & \multicolumn{2}{c}{\textbf{AVG}} \\
\cmidrule(lr){3-4} \cmidrule(lr){5-6} \cmidrule(lr){7-8} \cmidrule(lr){9-10}
 & & \scriptsize{\textbf{BLEURT}} & \scriptsize{\textbf{COMET}} & \scriptsize{\textbf{BLEURT}} & \scriptsize{\textbf{COMET}} & \scriptsize{\textbf{BLEURT}} & \scriptsize{\textbf{COMET}} & \scriptsize{\textbf{BLEURT}} & \scriptsize{\textbf{COMET}} \\
\midrule
\multicolumn{2}{l}{Llama2-7B-SFT} & 76.78 & 87.43 & 72.05 & 85.91 & 61.92 & 80.05 & 64.42 & 81.38 \\\midrule
\multicolumn{2}{l}{w/ Direct Trans.} & 76.70$_{\text{-0.08}}$ & 87.39$_{\text{-0.04}}$ & 71.75$_{\text{-0.30}}$ & 85.60$_{\text{-0.31}}$ & 62.99$_{\text{+1.07}}$ & 80.54$_{\text{+0.49}}$ & 65.24$_{\text{+0.82}}$ & 81.73$_{\text{+0.35}}$ \\

\multicolumn{2}{l}{\quad\quad w/ EAxE}  & 77.18$_{\text{+0.40}}$ & 87.67$_{\text{+0.24}}$ & 72.95$_{\text{+0.90}}$ & 86.64$_{\text{+0.73}}$ & 68.04$_{\text{+6.12}}$ & 83.51$_{\text{+3.46}}$ & 69.45$_{\text{+5.03}}$ & 84.24$_{\text{+2.86}}$ \\ \midrule
\multicolumn{2}{l}{w/ EAxT.} & 76.89$_{\text{+0.11}}$ & 87.48$_{\text{+0.05}}$ & 71.53$_{\text{-0.52}}$ & 85.47$_{\text{-0.44}}$ & 67.71$_{\text{+5.79}}$ & 82.88$_{\text{+2.83}}$ & 69.01$_{\text{+4.59}}$ & 83.60$_{\text{+2.22}}$ \\
\multicolumn{2}{l}{\quad\quad w/ EAxE} & 77.15$_{\text{+0.37}}$ & 87.62$_{\text{+0.19}}$ & 73.06$_{\text{+1.01}}$ & 86.74$_{\text{+0.83}}$ & 68.91$_{\text{+6.99}}$ & 83.96$_{\text{+3.91}}$ & 70.15$_{\text{+5.73}}$ & 84.60$_{\text{+3.22}}$ \\
\bottomrule
\end{tabular}
\caption{Ablation study evaluating English-Anchored x2x Translation and Evaluation mechanisms on the Llama2 model using the FLORES-200 testset. We labeled the performance delta of each combination with respect to the SFT baseline.}
\label{tab:ablation_overview}
\end{table*}

\begin{table}[htbp]
\small
\centering
\begin{tabular}{lcc}
\toprule
\textbf{Synthetic Strategy} & \textbf{BLEURT} & \textbf{COMET}\\
\midrule
Direct     & 68.04 & 83.51  \\
Pivot      & 68.58 & 83.35  \\
EAxT      & 68.91 & 83.96  \\
\midrule
\textbf{Metric for EAxE} &  & \\\midrule

Random     & 67.71 & 82.88  \\
PPL      & 67.78 & 82.85  \\
KIWI      & 68.71 & 83.86  \\
Direct RM & 67.97 & 83.39  \\
RM      & 68.91 & 83.96  \\

\bottomrule
\end{tabular}
\caption{The x2x performance on the FLORES-200 testset of optimized Llama2 with different data synthesis strategies and alternative metrics for preference construction.}
\label{tab:ablation_synthetic_metric}
\end{table}

We first investigate the effects of two key components: the English-Anchored x2x Translation~(EAxT)-based data synthesis strategy and the English-Anchored x2x Evaluation(EAxE)-driven data selection mechanism. For EAxT ablation, we substitute our method with direct translation outputs. When disabling the reward model for EAxE, we randomly select translation candidates and perform standard fine-tuning rather than preference optimization.

As shown in Table~\ref{tab:ablation_overview}, without applying any of our proposed methods, the improvements obtained from fine-tuning on directly synthesized data are quite limited. Each of our two proposed enhancements contributed significantly to translation performance improvement. In particular, the utilization of the reward model effectively mitigate the quality deficiencies in directly synthesized data, highlighting the necessity of data selection and cleaning for synthetic data.

Furthermore, we observe that performance improvements in en2x translation directions are also achieved through preference data constructed via the reward model.
This aligns with the emerging consensus that reinforcement learning yields better generalization compared to standard supervised fine-tuning~\cite{chu2025sft}.
We further validate this hypothesis in Section~\ref{sec:scaling}.

Table~\ref{tab:ablation_synthetic_metric} presents a comprehensive analysis of the influence of three distinct data synthesis methods on the resultant x2x translation performance metrics.
Generally speaking, all methods effectively construct preferences to enhance the model's x2x translation capabilities. Nevertheless, EAxT further elevates the model's performance ceiling.

We further consider available quality assessment metrics as alternatives to the reward model. The following baselines are evaluated:
\begin{compactitem}
    \item Random picking followed by fine-tuning.
    \item Translation model perplexity (PPL).
    \item {\small COMETKIWI-XL}~\cite{rei-etal-2023-scaling}, a model specifically designed for translation quality estimation without requiring reference translations.
\end{compactitem}

In addition, we explore using our reward model for direct evaluation of x2x translations (Direct RM), with the wondering whether its evaluation capabilities can transfer to x2x language pairs. Specifically, we directly provide the source text $x_{l_1}$ to the reward model instead of its English reference, computing the score as $s^i=r(x_{l_1},\bar{x}_{l_2}^i)$.

As shown in Table~\ref{tab:ablation_synthetic_metric}, PPL performs comparably to the random baseline, indicating that translation models cannot be directly used for evaluation without appropriate training to activate their assessment capabilities, e.g., through reward modeling. Our method slightly outperforms {\small COMETKIWI}, suggesting the potential of LLM-driven quality assessment, particularly given its independence from annotated translation evaluation data. Finally, we observe that the evaluation capabilities of our reward model can partially generalize to x2x language pairs, although this direct application is notably less effective than the proxy evaluation approach.

\begin{figure}[t]
    \centering
    \begin{subfigure}[b]{0.24\textwidth}
        \centering 
        \includegraphics[width=\textwidth]{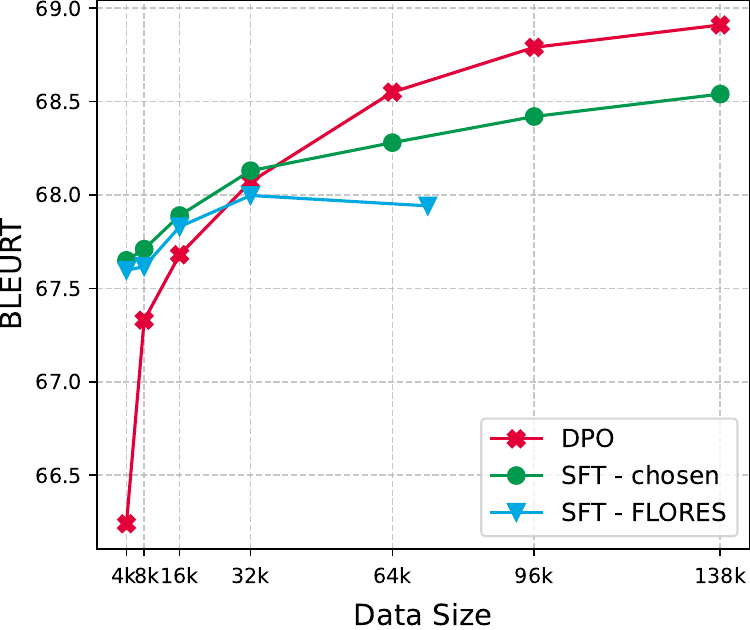}
        \caption{\textbf{x2x}}
        \label{fig:scaling_x2x}
    \end{subfigure}%
    \begin{subfigure}[b]{0.24\textwidth}
        \centering
        \includegraphics[width=\textwidth]{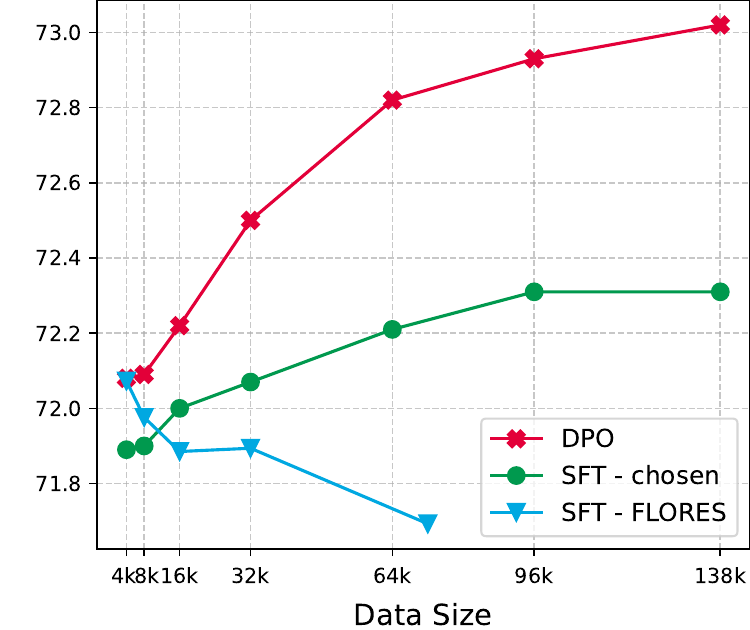}
        \caption{\textbf{en2x}}
        \label{fig:scaling_en2x}
    \end{subfigure}
    \vspace{-1ex}
    \caption{Performance on the FLORES-200 testset of each optimization algorithm scaling with data size.}
    \label{fig:scaling}
\end{figure}

\subsection{Scaling with Synthetic Data}
\label{sec:scaling}

This section highlights the advantages of synthetic data scaling, particularly comparing the translation improvements through preference optimization versus vanilla supervised fine-tuning across varying data scales, as well as their generalization disparities on unseen language pairs (en2x).
Specifically, we control the scale of preference data used for optimization, and for comparison, we fine-tune only on the chosen data from the preference data pairs.
For comprehensive evaluation, we additionally incorporated human-annotated data from the FLORES devset.

Figure~\ref{fig:scaling_x2x} illustrates the trend in translation performance on x2x language pairs using different optimization algorithms. Initially, DPO lags behind SFT on small-scale data. However, as the data size increases, DPO demonstrates continuous improvement, rapidly surpassing the SFT baselines and maintaining its advantage with further scaling. Although SFT trained on chosen data also improves with scale, its gains are comparatively modest.

In the en2x translation scenario shown in Figure~\ref{fig:scaling_en2x}, the performance advantage of DPO becomes even more pronounced, indicating superior generalization effects for unseen language pairs.

For FLORES data, constraints of data scale necessitate text reuse across different language pairs, introducing the risk of model overfitting. Consequently, the limitations in data diversity manifest as limited scalability with increased data size, and even slight performance degradation, particularly in en2x translation.

\section{Related Work}

\paragraph{LLM-Driven Data Synthesis}

LLM-driven synthetic data generation has emerged as a promising alternative to traditional human-dependent data collection, demonstrating significant potential across various applications. In the context of NLP tasks, LLMs have been extensively integrated into data generation pipelines, encompassing areas such as question answering~\cite{li-callison-burch-2023-paxqa}, text classification~\cite{li-etal-2023-synthetic}, and general capabilities~\cite{huang-etal-2023-large}. These efforts have underscored the importance of curation, evaluation, and quality control of synthetic data. Additionally, the paradigm of utilizing synthetic data to replace human annotation has found applications in domain-specific tasks~\cite{tang2023does} and multimodal fields~\cite{liu2024improved}.

\paragraph{Many-To-Many Translation}

Developing many-to-many translation capabilities for machine translation models is a challenging task. Previous work based on neural machine translation (NMT) has explored a range of techniques, such as introducing representation alignment~\cite{pan2021contrastive} or achieving flexible combinations of language pairs through shared encoders and decoders~\cite{yuan-etal-2023-lego} or Mixture-of-Experts~\cite{fan2020beyond, nllb2022flores} architectures. Nevertheless, large-scale many-to-many translation datasets obtained through mining remain essential~\cite{yuan-etal-2023-lego, fan2020beyond, nllb2022flores}.

For LLMs, prior research has demonstrated that multilingual capabilities exhibit inherent imbalances between English and non-English languages \cite{yuan-etal-2024-vocabulary}.
This disparity is primarily attributed to the uneven language distribution in pretraining data.
Consequently, existing works aim to address the deficiencies of LLMs in non-English languages and enhance many-to-many translation capabilities through large-scale continued pre-training~\cite{lu-etal-2024-llamax, zheng2025asymmetric}. These efforts typically require substantial monolingual and parallel data across many languages.

In contrast, we focus on post-training of LLMs. Our findings suggest that even models enhanced for multilingual capabilities, such as Tower (which undergoes continued pretraining) or Qwen (which uses more diverse multilingual data), may still amplify disparities between English and non-English capabilities without delicated adjustments. Our research complements existing approaches by fully activating LLMs' many-to-many translation capabilities within the framework of their foundational competencies.

\section{Conclusion}

In this work, we presented a novel approach to enhance x2x translation capabilities in large language models without requiring extensive non-English parallel data.
By leveraging English parallel corpora and the inherent en2x strengths of LLMs, we proposed a synthesis and evaluation framework to enhance x2x translation capabilities. 
This method not only boosts x2x translation quality but also unexpectedly enhances en2x performance, indicating robust generalization across languages.
These findings suggest promising directions for future research in multilingual translation systems that can operate effectively across all language pairs beyond English. By reducing the reliance on scarce non-English parallel data, our approach offers a practical solution to the challenges of building truly omnidirectional translation systems.

\section*{Limitations}

Our experiments have investigated the feasibility of building many-to-many translation capabilities among mainstream languages. However, we have not yet explored the reaction of our approach when applied to low-resource languages. In particular, the implementation of our method may face significant challenges due to the scarcity of English-centric parallel data for low-resource languages. This data deficiency presents a substantial obstacle to the direct application of our approach in these linguistic contexts.

One potential solution to address this limitation would be to consider synthesizing parallel data from English to low-resource languages. Nevertheless, this strategy might be constrained by the model's inherent translation capabilities between English and these low-resource languages. The quality of synthetic data would inevitably depend on the model's proficiency in translating between these language pairs, which may be suboptimal given the limited training resources available for such languages.

Furthermore, the linguistic diversity and structural differences characteristic of many low-resource languages may introduce additional complexities that our current methodology does not explicitly account for. Future work should systematically investigate adaptations of our approach to accommodate the unique challenges presented by low-resource language translation scenarios.

\section*{Acknowledgments}

We would like to thank the anonymous reviewers for their insightful comments. Shujian Huang and Shanbo Cheng are the corresponding authors. This work is supported by National Science Foundation of China (No. 62376116, 62176120), research project of Nanjing University-China Mobile Joint Institute (NJ20250038), the Fundamental Research Funds for the Central Universities (No. 2024300507, 2025300390).

\bibliography{anthology,custom}

\begin{thebibliography}{30}
\providecommand{\natexlab}[1]{#1}

\bibitem[{AI(2023)}]{OpenAI2023}
Open AI. 2023.
\newblock Chatml documentation.
\newblock \url{https://github.com/openai/openai-python/blob/release-v0.28.1/chatml.md}.
\newblock Accessed: 2023-10-01.

\bibitem[{Alves et~al.(2024)Alves, Pombal, Guerreiro, Martins, Alves, Farajian, Peters, Rei, Fernandes, Agrawal et~al.}]{alves2024tower}
Duarte~M Alves, Jos{\'e} Pombal, Nuno~M Guerreiro, Pedro~H Martins, Jo{\~a}o Alves, Amin Farajian, Ben Peters, Ricardo Rei, Patrick Fernandes, Sweta Agrawal, and 1 others. 2024.
\newblock Tower: An open multilingual large language model for translation-related tasks.
\newblock \emph{arXiv preprint arXiv:2402.17733}.

\bibitem[{Chu et~al.(2025)Chu, Zhai, Yang, Tong, Xie, Schuurmans, Le, Levine, and Ma}]{chu2025sft}
Tianzhe Chu, Yuexiang Zhai, Jihan Yang, Shengbang Tong, Saining Xie, Dale Schuurmans, Quoc~V Le, Sergey Levine, and Yi~Ma. 2025.
\newblock Sft memorizes, rl generalizes: A comparative study of foundation model post-training.
\newblock \emph{arXiv preprint arXiv:2501.17161}.

\bibitem[{Costa-juss{\`a} et~al.(2022)Costa-juss{\`a}, Cross, {\c{C}}elebi, Elbayad, Heafield, Heffernan, Kalbassi, Lam, Licht, Maillard et~al.}]{nllb2022flores}
Marta~R Costa-juss{\`a}, James Cross, Onur {\c{C}}elebi, Maha Elbayad, Kenneth Heafield, Kevin Heffernan, Elahe Kalbassi, Janice Lam, Daniel Licht, Jean Maillard, and 1 others. 2022.
\newblock \href {https://arxiv.org/abs/2207.04672} {No language left behind: Scaling human-centered machine translation}.
\newblock \emph{ArXiv preprint}, abs/2207.04672.

\bibitem[{Fan et~al.(2020)Fan, Bhosale, Schwenk, Ma, El-Kishky, Goyal, Baines, Celebi, Wenzek, Chaudhary, Goyal, Birch, Liptchinsky, Edunov, Grave, Auli, and Joulin}]{fan2020beyond}
Angela Fan, Shruti Bhosale, Holger Schwenk, Zhiyi Ma, Ahmed El-Kishky, Siddharth Goyal, Mandeep Baines, Onur Celebi, Guillaume Wenzek, Vishrav Chaudhary, Naman Goyal, Tom Birch, Vitaliy Liptchinsky, Sergey Edunov, Edouard Grave, Michael Auli, and Armand Joulin. 2020.
\newblock Beyond english-centric multilingual machine translation.
\newblock \emph{arXiv preprint}.

\bibitem[{Huang et~al.(2023)Huang, Gu, Hou, Wu, Wang, Yu, and Han}]{huang-etal-2023-large}
Jiaxin Huang, Shixiang Gu, Le~Hou, Yuexin Wu, Xuezhi Wang, Hongkun Yu, and Jiawei Han. 2023.
\newblock \href {https://doi.org/10.18653/v1/2023.emnlp-main.67} {Large language models can self-improve}.
\newblock In \emph{Proceedings of the 2023 Conference on Empirical Methods in Natural Language Processing}, pages 1051--1068, Singapore. Association for Computational Linguistics.

\bibitem[{Kocmi et~al.(2022)Kocmi, Bawden, Bojar, Dvorkovich, Federmann, Fishel, Gowda, Graham, Grundkiewicz, Haddow, Knowles, Koehn, Monz, Morishita, Nagata, Nakazawa, Nov{\'a}k, Popel, and Popovi{\'c}}]{kocmi-etal-2022-findings}
Tom Kocmi, Rachel Bawden, Ond{\v{r}}ej Bojar, Anton Dvorkovich, Christian Federmann, Mark Fishel, Thamme Gowda, Yvette Graham, Roman Grundkiewicz, Barry Haddow, Rebecca Knowles, Philipp Koehn, Christof Monz, Makoto Morishita, Masaaki Nagata, Toshiaki Nakazawa, Michal Nov{\'a}k, Martin Popel, and Maja Popovi{\'c}. 2022.
\newblock \href {https://aclanthology.org/2022.wmt-1.1/} {Findings of the 2022 conference on machine translation ({WMT}22)}.
\newblock In \emph{Proceedings of the Seventh Conference on Machine Translation (WMT)}, pages 1--45, Abu Dhabi, United Arab Emirates (Hybrid). Association for Computational Linguistics.

\bibitem[{Kocmi and Federmann(2023)}]{kocmi-federmann-2023-large}
Tom Kocmi and Christian Federmann. 2023.
\newblock \href {https://aclanthology.org/2023.eamt-1.19/} {Large language models are state-of-the-art evaluators of translation quality}.
\newblock In \emph{Proceedings of the 24th Annual Conference of the European Association for Machine Translation}, pages 193--203, Tampere, Finland. European Association for Machine Translation.

\bibitem[{Li and Callison-Burch(2023)}]{li-callison-burch-2023-paxqa}
Bryan Li and Chris Callison-Burch. 2023.
\newblock \href {https://doi.org/10.18653/v1/2023.findings-emnlp.32} {{PAXQA}: Generating cross-lingual question answering examples at training scale}.
\newblock In \emph{Findings of the Association for Computational Linguistics: EMNLP 2023}, pages 439--454, Singapore. Association for Computational Linguistics.

\bibitem[{Li et~al.(2023)Li, Zhu, Lu, and Yin}]{li-etal-2023-synthetic}
Zhuoyan Li, Hangxiao Zhu, Zhuoran Lu, and Ming Yin. 2023.
\newblock \href {https://doi.org/10.18653/v1/2023.emnlp-main.647} {Synthetic data generation with large language models for text classification: Potential and limitations}.
\newblock In \emph{Proceedings of the 2023 Conference on Empirical Methods in Natural Language Processing}, pages 10443--10461, Singapore. Association for Computational Linguistics.

\bibitem[{Liu et~al.(2024{\natexlab{a}})Liu, Li, Li, and Lee}]{liu2024improved}
Haotian Liu, Chunyuan Li, Yuheng Li, and Yong~Jae Lee. 2024{\natexlab{a}}.
\newblock Improved baselines with visual instruction tuning.
\newblock In \emph{Proceedings of the IEEE/CVF Conference on Computer Vision and Pattern Recognition}, pages 26296--26306.

\bibitem[{Liu et~al.(2024{\natexlab{b}})Liu, Wei, Liu, Si, Zhang, Rao, Zheng, Peng, Yang, Zhou et~al.}]{liu2024best}
Ruibo Liu, Jerry Wei, Fangyu Liu, Chenglei Si, Yanzhe Zhang, Jinmeng Rao, Steven Zheng, Daiyi Peng, Diyi Yang, Denny Zhou, and 1 others. 2024{\natexlab{b}}.
\newblock Best practices and lessons learned on synthetic data.
\newblock \emph{arXiv preprint arXiv:2404.07503}.

\bibitem[{Long et~al.(2024)Long, Wang, Xiao, Zhao, Ding, Chen, and Wang}]{long-etal-2024-llms}
Lin Long, Rui Wang, Ruixuan Xiao, Junbo Zhao, Xiao Ding, Gang Chen, and Haobo Wang. 2024.
\newblock \href {https://doi.org/10.18653/v1/2024.findings-acl.658} {On {LLM}s-driven synthetic data generation, curation, and evaluation: A survey}.
\newblock In \emph{Findings of the Association for Computational Linguistics: ACL 2024}, pages 11065--11082, Bangkok, Thailand. Association for Computational Linguistics.

\bibitem[{Loshchilov and Hutter(2019)}]{loshchilov2019decoupledweightdecayregularization}
Ilya Loshchilov and Frank Hutter. 2019.
\newblock \href {https://arxiv.org/abs/1711.05101} {Decoupled weight decay regularization}.
\newblock \emph{Preprint}, arXiv:1711.05101.

\bibitem[{Lu et~al.(2024)Lu, Zhu, Li, Qiao, and Yuan}]{lu-etal-2024-llamax}
Yinquan Lu, Wenhao Zhu, Lei Li, Yu~Qiao, and Fei Yuan. 2024.
\newblock \href {https://doi.org/10.18653/v1/2024.findings-emnlp.631} {{LL}a{MAX}: Scaling linguistic horizons of {LLM} by enhancing translation capabilities beyond 100 languages}.
\newblock In \emph{Findings of the Association for Computational Linguistics: EMNLP 2024}, pages 10748--10772, Miami, Florida, USA. Association for Computational Linguistics.

\bibitem[{Pan et~al.(2021)Pan, Wang, Wu, and Li}]{pan2021contrastive}
Xiao Pan, Mingxuan Wang, Liwei Wu, and Lei Li. 2021.
\newblock Contrastive learning for many-to-many multilingual neural machine translation.
\newblock \emph{arXiv preprint arXiv:2105.09501}.

\bibitem[{Qwen et~al.(2025)Qwen, :, Yang, Yang, Zhang, Hui, Zheng, Yu, Li, Liu, Huang, Wei, Lin, Yang, Tu, Zhang, Yang, Yang, Zhou, Lin, Dang, Lu, Bao, Yang, Yu, Li, Xue, Zhang, Zhu, Men, Lin, Li, Tang, Xia, Ren, Ren, Fan, Su, Zhang, Wan, Liu, Cui, Zhang, and Qiu}]{qwen2025qwen25technicalreport}
Qwen, :, An~Yang, Baosong Yang, Beichen Zhang, Binyuan Hui, Bo~Zheng, Bowen Yu, Chengyuan Li, Dayiheng Liu, Fei Huang, Haoran Wei, Huan Lin, Jian Yang, Jianhong Tu, Jianwei Zhang, Jianxin Yang, Jiaxi Yang, Jingren Zhou, and 25 others. 2025.
\newblock \href {https://arxiv.org/abs/2412.15115} {Qwen2.5 technical report}.
\newblock \emph{Preprint}, arXiv:2412.15115.

\bibitem[{Rafailov et~al.(2023)Rafailov, Sharma, Mitchell, Manning, Ermon, and Finn}]{rafailov2023direct}
Rafael Rafailov, Archit Sharma, Eric Mitchell, Christopher~D Manning, Stefano Ermon, and Chelsea Finn. 2023.
\newblock Direct preference optimization: Your language model is secretly a reward model.
\newblock \emph{Advances in Neural Information Processing Systems}, 36:53728--53741.

\bibitem[{Rei et~al.(2022)Rei, C.~de Souza, Alves, Zerva, Farinha, Glushkova, Lavie, Coheur, and Martins}]{rei-etal-2022-comet}
Ricardo Rei, Jos{\'e}~G. C.~de Souza, Duarte Alves, Chrysoula Zerva, Ana~C Farinha, Taisiya Glushkova, Alon Lavie, Luisa Coheur, and Andr{\'e} F.~T. Martins. 2022.
\newblock \href {https://aclanthology.org/2022.wmt-1.52/} {{COMET}-22: Unbabel-{IST} 2022 submission for the metrics shared task}.
\newblock In \emph{Proceedings of the Seventh Conference on Machine Translation (WMT)}, pages 578--585, Abu Dhabi, United Arab Emirates (Hybrid). Association for Computational Linguistics.

\bibitem[{Rei et~al.(2023)Rei, Guerreiro, Pombal, van Stigt, Treviso, Coheur, C.~de Souza, and Martins}]{rei-etal-2023-scaling}
Ricardo Rei, Nuno~M. Guerreiro, Jos{\~A}{\textcopyright} Pombal, Daan van Stigt, Marcos Treviso, Luisa Coheur, Jos{\'e}~G. C.~de Souza, and Andr{\'e} Martins. 2023.
\newblock \href {https://doi.org/10.18653/v1/2023.wmt-1.73} {Scaling up {C}omet{K}iwi: Unbabel-{IST} 2023 submission for the quality estimation shared task}.
\newblock In \emph{Proceedings of the Eighth Conference on Machine Translation}, pages 841--848, Singapore. Association for Computational Linguistics.

\bibitem[{Seddik et~al.(2024)Seddik, Chen, Hayou, Youssef, and Debbah}]{seddik2024bad}
Mohamed El~Amine Seddik, Suei-Wen Chen, Soufiane Hayou, Pierre Youssef, and Merouane Debbah. 2024.
\newblock How bad is training on synthetic data? a statistical analysis of language model collapse.
\newblock \emph{arXiv preprint arXiv:2404.05090}.

\bibitem[{Sellam et~al.(2020)Sellam, Das, and Parikh}]{sellam-etal-2020-bleurt}
Thibault Sellam, Dipanjan Das, and Ankur Parikh. 2020.
\newblock \href {https://doi.org/10.18653/v1/2020.acl-main.704} {{BLEURT}: Learning robust metrics for text generation}.
\newblock In \emph{Proceedings of the 58th Annual Meeting of the Association for Computational Linguistics}, pages 7881--7892, Online. Association for Computational Linguistics.

\bibitem[{Tang et~al.(2023)Tang, Han, Jiang, and Hu}]{tang2023does}
Ruixiang Tang, Xiaotian Han, Xiaoqian Jiang, and Xia Hu. 2023.
\newblock Does synthetic data generation of llms help clinical text mining?
\newblock \emph{arXiv preprint arXiv:2303.04360}.

\bibitem[{Touvron et~al.(2023)Touvron, Martin, Stone, Albert, Almahairi, Babaei, Bashlykov, Batra, Bhargava, Bhosale et~al.}]{touvron2023llama}
Hugo Touvron, Louis Martin, Kevin Stone, Peter Albert, Amjad Almahairi, Yasmine Babaei, Nikolay Bashlykov, Soumya Batra, Prajjwal Bhargava, Shruti Bhosale, and 1 others. 2023.
\newblock Llama 2: Open foundation and fine-tuned chat models.
\newblock \emph{arXiv preprint arXiv:2307.09288}.

\bibitem[{Xu et~al.(2023)Xu, Kim, Sharaf, and Awadalla}]{xu2023paradigm}
Haoran Xu, Young~Jin Kim, Amr Sharaf, and Hany~Hassan Awadalla. 2023.
\newblock A paradigm shift in machine translation: Boosting translation performance of large language models.
\newblock \emph{arXiv preprint arXiv:2309.11674}.

\bibitem[{Yu et~al.(2023)Yu, Zhuang, Zhang, Meng, Ratner, Krishna, Shen, and Zhang}]{yu2023large}
Yue Yu, Yuchen Zhuang, Jieyu Zhang, Yu~Meng, Alexander~J Ratner, Ranjay Krishna, Jiaming Shen, and Chao Zhang. 2023.
\newblock Large language model as attributed training data generator: A tale of diversity and bias.
\newblock \emph{Advances in Neural Information Processing Systems}, 36:55734--55784.

\bibitem[{Yuan et~al.(2023)Yuan, Lu, Zhu, Kong, Li, Qiao, and Xu}]{yuan-etal-2023-lego}
Fei Yuan, Yinquan Lu, Wenhao Zhu, Lingpeng Kong, Lei Li, Yu~Qiao, and Jingjing Xu. 2023.
\newblock \href {https://doi.org/10.18653/v1/2023.findings-acl.731} {{L}ego-{MT}: Learning detachable models for massively multilingual machine translation}.
\newblock In \emph{Findings of the Association for Computational Linguistics: ACL 2023}, pages 11518--11533, Toronto, Canada. Association for Computational Linguistics.

\bibitem[{Yuan et~al.(2024)Yuan, Yuan, Wu, and Li}]{yuan-etal-2024-vocabulary}
Fei Yuan, Shuai Yuan, Zhiyong Wu, and Lei Li. 2024.
\newblock \href {https://doi.org/10.18653/v1/2024.findings-acl.721} {How vocabulary sharing facilitates multilingualism in {LL}a{MA}?}
\newblock In \emph{Findings of the Association for Computational Linguistics: ACL 2024}, pages 12111--12130, Bangkok, Thailand. Association for Computational Linguistics.

\bibitem[{Zhang and Toral(2019)}]{zhang-toral-2019-effect}
Mike Zhang and Antonio Toral. 2019.
\newblock \href {https://doi.org/10.18653/v1/W19-5208} {The effect of translationese in machine translation test sets}.
\newblock In \emph{Proceedings of the Fourth Conference on Machine Translation (Volume 1: Research Papers)}, pages 73--81, Florence, Italy. Association for Computational Linguistics.

\bibitem[{Zheng et~al.(2025)Zheng, Wen, Bao, Guo, and Huang}]{zheng2025asymmetric}
Tong Zheng, Yan Wen, Huiwen Bao, Junfeng Guo, and Heng Huang. 2025.
\newblock Asymmetric conflict and synergy in post-training for llm-based multilingual machine translation.
\newblock \emph{arXiv preprint arXiv:2502.11223}.

\end{thebibliography}

\appendix

\section{Implementation Details}

\begin{figure}[tb]
    \centering
    \includegraphics[width=0.3\textwidth]{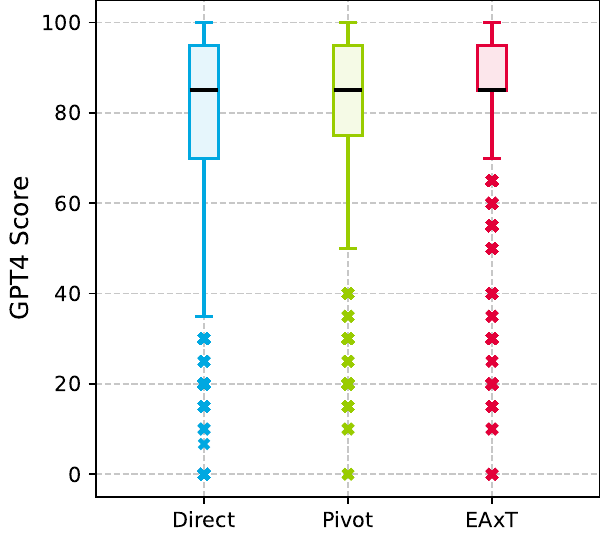}
    \vspace{-1ex}
   \caption{Box plot of GPT4 quality scores. Indicate the distribution of boxes, whiskers, and outliers for each synthetic method.}
    \label{fig:gpt_score}
\end{figure}

\label{sec:details}

\paragraph{Prompt Template}

For data generation and model evaluation, we use the prompt as illustrated in Figure 1 for sampling. For training, we refer to \citet{alves2024tower} and adopt diverse translation prompts, while removing the few-shot setting and retaining only the zero-shot setting. A total of 21 prompt variants are used in our training, which are accessible in our code base.

\paragraph{Sampling Strategies}
For the en2x translation generation involved in reward modeling of Section~\ref{sec:reward_model}, we employ a standard sampling strategy with a temperature of 1 and top-p of 1. For the generation of x2x translation data, we control the temperature at 0.9 and top-p at 0.6 to mitigate the risk of quality degradation. For each input, we sample 4 candidates.

\paragraph{Preference Data Filtering}

Our method employs Reward Modeling and Direct Preference Optimization, both of which are built on high-quality preference data. To improve the accuracy and consistency of this data, we use a straightforward heuristic to filter out pairs with small score margins between the chosen and rejected responses.

For the final preference data used in x2x optimization, we do not report the specific margin value, as it varies based on the score distribution of the specific reward model. Overall, we filtered out approximately 80\% of the data.

In Reward Modeling, where preference data is scored using BLEURT, we apply a margin of 20. Additionally, we require a minimum BLEURT score of 70 for the chosen data. As a result, we filtered about 50\% of the data from the TowerBlocks en-x 15k dataset.

\paragraph{Training Setups}
We list the training hyperparameters involved in each stage in Table~\ref{tab:hparams}.
All training was conducted using 16 Ascend 910B NPUs, equipped with bf16 mixed precision training, and utilizes DeepSpeed ZeRO-3 for sharding. Following the setup of TowerInstruct~\cite{alves2024tower}, we use the \texttt{chatml} template~\cite{OpenAI2023} during both training and inference, as well as instruction diversity, providing multiple zero-shot instruction templates for the translation task.

\begin{table*}[t]
\begin{threeparttable}[b]
\centering
\begin{tabular}{lllllll}
\toprule
 & \small{SFT on en-x} & \small{SFT on FLORES} & \small{SFT on chosen} & \small{DPO} & \small{RM} \\
\midrule
Global batch size & 128 & 128 & 128 & 64  & 64 \\
Train epoch & 1 & 1 & 1 & 1 & 1 \\
Learning rate & 7e-6 & 1e-6 & 4e-6 & 2e-7 & 4e-6 \\
Learning rate Decay & cosine & cosine & cosine & cosine & cosine \\
Warmup ratio & 0.1 & 0.1 & 0.1 & 0.1 & 0.1 \\
Optimizer & AdamW\tnote{$\dagger$} & AdamW\tnote{$\dagger$} & AdamW\tnote{$\dagger$} & AdamW\tnote{$\dagger$} & AdamW\tnote{$\dagger$}  \\
Weight Decay & 0 & 0 & 0 & 0 & 0 \\
Adam $\beta_1$ & 0.9 & 0.9 & 0.9 & 0.9 & 0.9 \\
Adam $\beta_2$ & 0.999 & 0.999 & 0.999 & 0.999 & 0.999 \\
Adam $\epsilon$ & 0 & 0 & 0 & 0 & 0  \\
Max Seq Len & 2048 & 2048 & 2048 & 2048 & 2048 \\
DPO $\beta$ & - & - & - & 0.4 (0.2 for Llama) & - \\
SFT coefficient\tnote{$\dagger\dagger$} & - & - & - & 2.0 & - \\
\bottomrule
\end{tabular}
\begin{tablenotes}
\footnotesize 
\item[$\dagger$] \citealp{loshchilov2019decoupledweightdecayregularization}.
\item[$\dagger\dagger$] The supervised fine-tuning loss coefficient in DPO training.
\end{tablenotes}
\caption{Hyperparameter configuration for SFT, DPO and RM training.}
\label{tab:hparams}
\end{threeparttable}
\end{table*}

\section{GPT-Based Metric Results}
\label{sec:gpt_score}

Figure~\ref{fig:gpt_score} illustrates the distribution of translation quality across the synthetic data methods.
Comparing the three, the box position of EAxT suggests potential for higher - scoring translations. Pivot and Direct has a wider spread and outliers.
Our method, despite having a narrow box and whiskers, still exhibits a substantial number of low-scoring outliers, suggesting that while it generally performs consistently, it cannot prevent the generation of poor-quality translation results without our evaluation curation.

\section{Results in Individual Languages}
\label{sec:results_in_langs}

\begin{figure*}[htb]
    \centering
    \includegraphics[width=\textwidth]{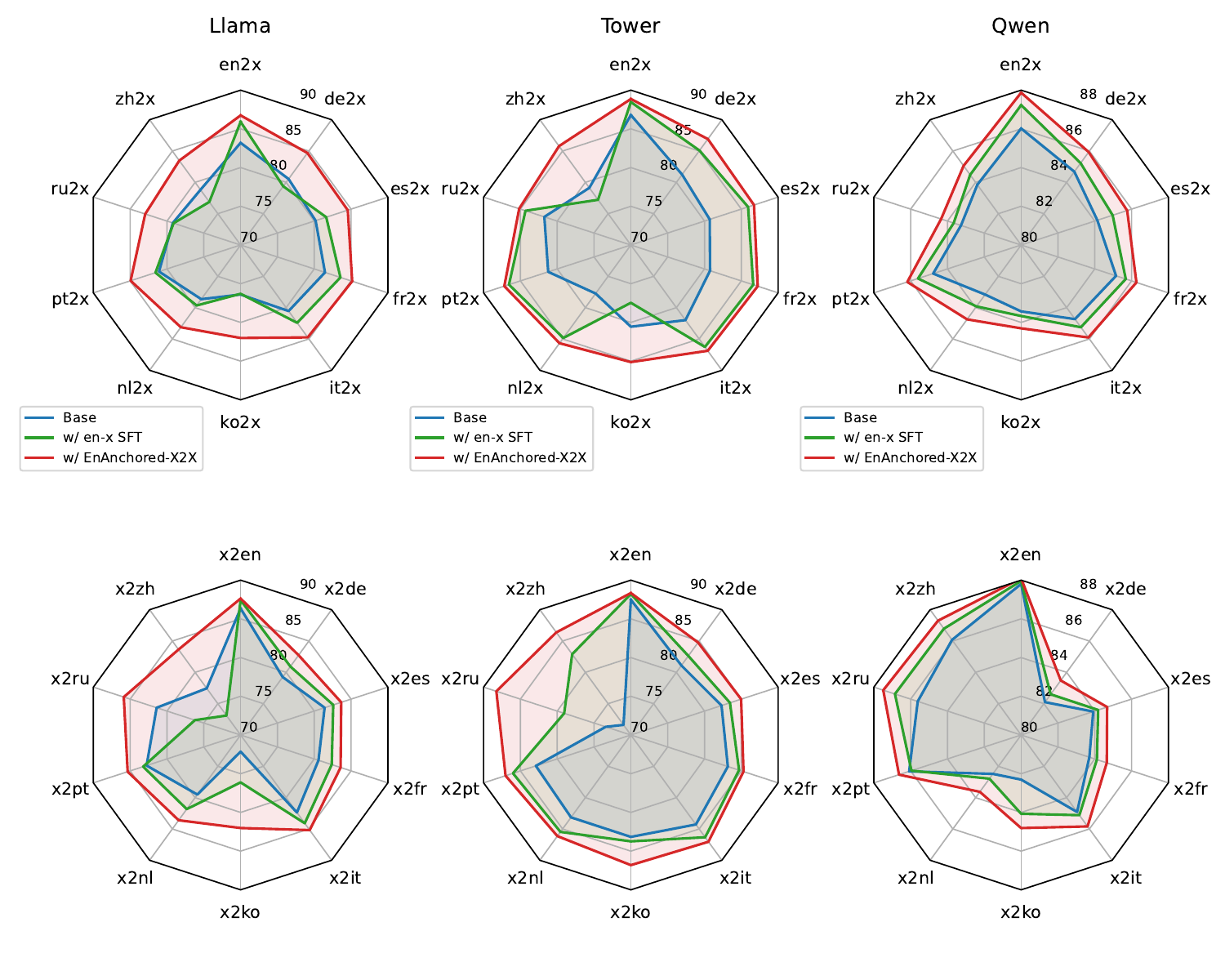}
    \vspace{-3ex}
   \caption{COMET22 performance on FLORES-200 testset with each language as source or target.}
    \label{fig:comet_in_langs}
\end{figure*}

\begin{figure*}[htb]
    \centering
    \includegraphics[width=\textwidth]{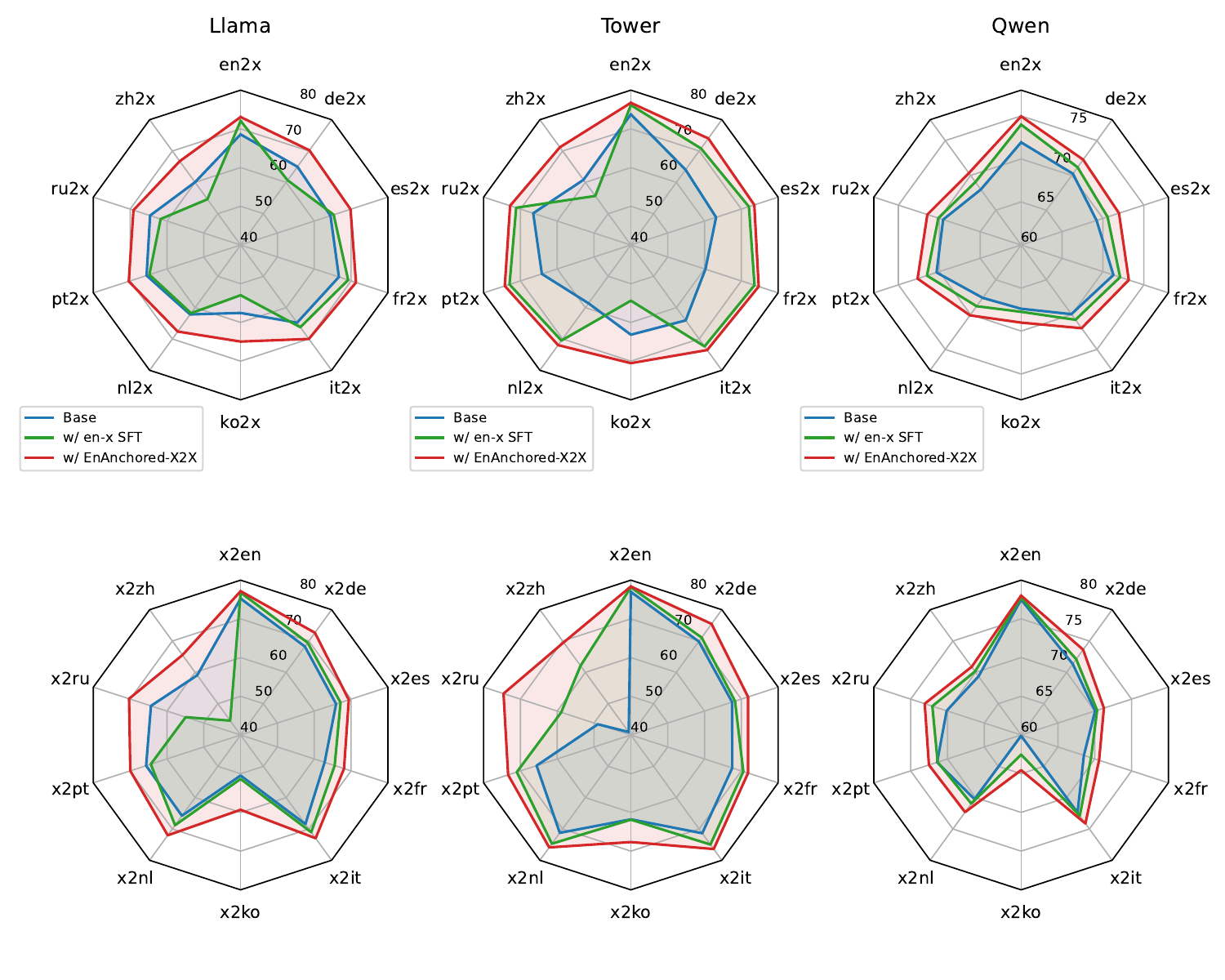}
    \vspace{-3ex}
   \caption{BLEURT performance on FLORES-200 testset with each language as source or target.}
    \label{fig:bleurt_in_langs}
\end{figure*}

In Figures~\ref{fig:comet_in_langs} and \ref{fig:bleurt_in_langs}, we respectively delineate the COMET and BLEURT performance across languages, presenting the performance improvements of en-x SFT, and our x2x optimization.

\end{document}